\documentclass{MIPRO}
\usepackage{comment}
\usepackage{amsmath,amssymb,amsfonts}
\usepackage{algorithmic}
\usepackage{graphicx}
\graphicspath{{images/}}
\usepackage{textcomp}
\usepackage{wrapfig}
\usepackage{longtable}

\usepackage[utf8]{inputenc}
\usepackage[T1]{fontenc}
\usepackage[english]{babel}
\usepackage{textcomp}
\usepackage{epstopdf}
\usepackage{subcaption}
\usepackage{tabularx}
\usepackage{gensymb}
\usepackage[hyphens]{url}

\usepackage{todonotes}
\usepackage{array}
\usepackage{booktabs}
\usepackage{xcolor, soul}

\usepackage{cite}
\usepackage{amsmath,amssymb,amsfonts}
\usepackage{algorithmic}
\usepackage{graphicx}
\usepackage{textcomp}
\usepackage{xcolor}
\usepackage[T1]{fontenc}
\usepackage{flushend}
\usepackage{multirow}
\usepackage{multicol}
\usepackage{array}
\usepackage{url}
\usepackage{fancyhdr}

\fancypagestyle{withfooter}{
  
  \fancyfoot[C]{\footnotesize Accepted to the IEEE IROS workshop on Autonomous Robotic Systems in Aquaculture: Research Challenges and Industry Needs}
}
\begin{document}

\thispagestyle{withfooter}
\pagestyle{withfooter}
\title{Autonomous Visual Fish Pen Inspections for Estimating the State of Biofouling Buildup Using ROV - Extended Abstract}

\author{
\IEEEauthorblockN{
Matej Fabijanić \IEEEauthorrefmark{1}, 
Nadir Kapetanović \IEEEauthorrefmark{1}, 
Nikola Mišković \IEEEauthorrefmark{1}
}

\IEEEauthorblockA{\IEEEauthorrefmark{1} 
University of Zagreb Faculty of Electrical Engineering and Computing, Zagreb, Croatia}



\thanks{
Research work presented in this article has been supported by 
the European Union under Grant Agreement Number 101060395 - MONUSEN project (MONtenegrin center for Underwater SEnsor Networks. Views and opinions expressed are however those of the author(s) only and do not necessarily reflect those of the European Union or the European Research Executive Agency (REA). Neither the European Union nor the granting authority can be held responsible for them;
the “Razvoj autonomnog besposadnog višenamjenskog broda” project (KK.01.2.1.02.0342) co-financed by the European Union from the European Regional Development Fund within the Operational Program “Competitiveness and Cohesion 2014-2020”;
the Croatian National Recovery and Resilience funded project Smart Blue Tourism, G.A. No. NPOO.C1.6.R1-I2.01-V3.0007;
the content of the publication is the sole responsibility of the project partner UNIZG-FER; and by ONR Robot Aided Diver Navigation in Mapped Environments - ROADMAP project under Grant Agreement No. N000142112274}
}

\maketitle
\thispagestyle{withfooter}
\pagestyle{withfooter}
\begin{abstract}
The process of fish cage inspections, which is a necessary maintenance task at any fish farm, be it small-scale or industrial, is a task that has the potential to be fully automated. Replacing trained divers who perform regular inspections with autonomous marine vehicles would lower the costs of manpower and remove the risks associated with humans performing underwater inspections. Achieving such a level of autonomy implies developing an image processing algorithm that is capable of estimating the state of biofouling buildup. The aim of this work is to propose a complete solution for automating the said inspection process; from developing an autonomous control algorithm for an ROV, to automatically segmenting images of fish cages, and accurately estimating the state of biofouling. The first part is achieved by modifying a commercially available ROV with an acoustic SBL positioning system and developing a closed-loop control system. The second part is realized by implementing a proposed biofouling estimation framework, which relies on AI to perform image segmentation, and by processing images using established computer vision methods to obtain a rough estimate of the distance of the ROV from the fish cage. This also involved developing a labeling tool in order to create a dataset of images for the neural network performing the semantic segmentation to be trained on. The experimental results show the viability of using an ROV fitted with an acoustic transponder for autonomous missions, and demonstrate the biofouling estimation framework’s ability to provide accurate assessments, alongside satisfactory distance estimation capabilities. In conclusion, the achieved biofouling estimation accuracy showcases clear potential for use in the aquaculture industry.
\end{abstract}

\renewcommand\IEEEkeywordsname{Keywords}

\section{Introduction} \label{sec: Introduction}
There has been a growing acknowledgment of the crucial role played by small-scale fisheries and industrial aquaculture in ensuring global food security and nutrition in the 21st century.
Global aquaculture production has shown a rising trend over the last 30~years, with around 80 million tonnes of seafood produced in 2020 \cite{Aquaculutre}.
Aquaculture is known to be highly labor-intensive, requiring significant human involvement in various tasks such as feeding, cleaning and processing as opposed to the envisioned streamlined machine-supported industrial agriculture.
The emergence of autonomous robots presents an opportunity to supplement and enhance these labor-intensive operations.
By integrating autonomous robots, the aquaculture sector can achieve higher efficiency, reduce operational costs, and ensure sustainable fishing practices for the future.

\begin{figure}[ht]
\includegraphics[width=\linewidth]{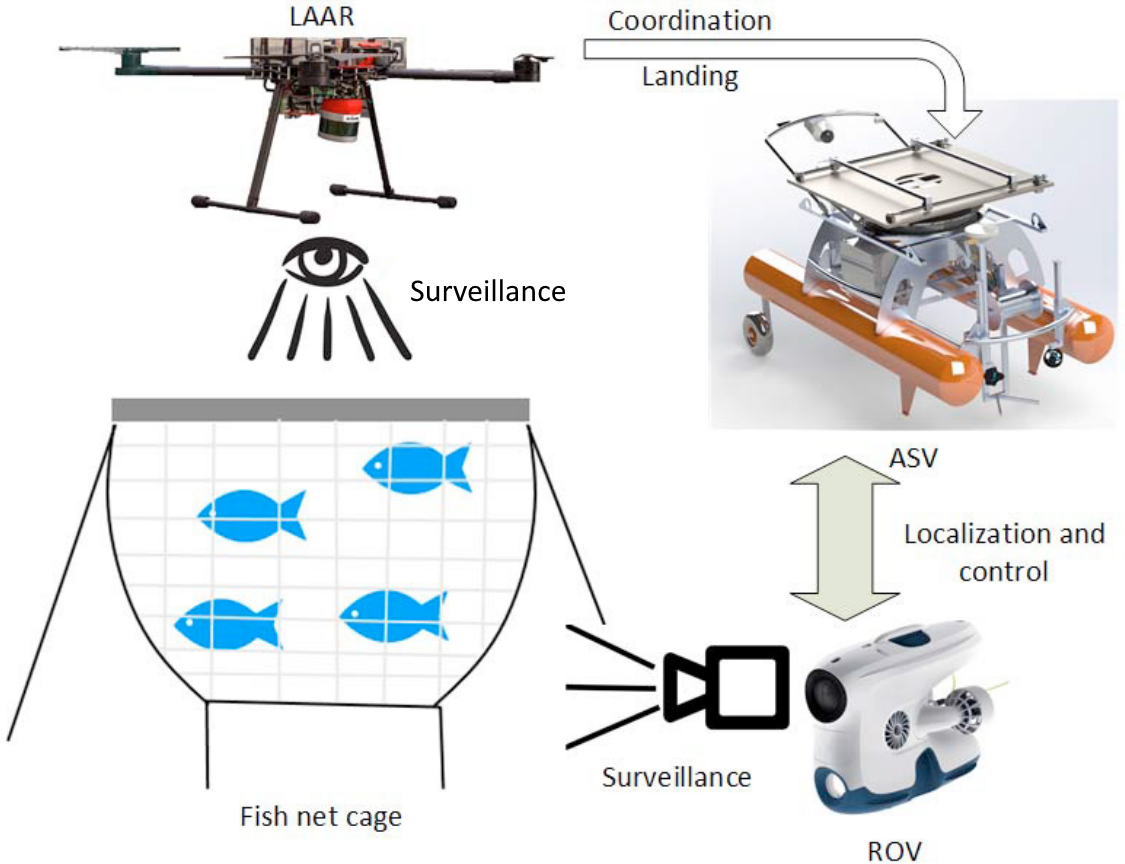}
\caption{A schematic of the proposed HEKTOR underwater inspection solution.}
\label{fig:hektorSchema}
\end{figure}

Previous research in this field often deals with only one specific aspect of aquaculture activities, like net damage detection in \cite{holeInspection1,holeInspection2}.
Other work such as that by Duda et al. \cite{poseEstimation} uses computer vision techniques to achieve ROV pose estimation and briefly touches on biological buildup on the net, but only mentions it as a potential problem for pose estimation.
More work by Livanos et al. \cite{movementButNoImageProc} discusses enhancing ROV autonomy level through intelligent navigation, but does not showcase a use for any specific fishery maintenance task.
Work by \mbox{Qiu et al. \cite{biofoulingOnly}} examines the estimation of built-up biofouling using image processing, but only briefly mentions using an ROV to capture footage needed for research.
To the best of the authors' knowledge, there is no literature on the development of a complete autonomous fish cage inspection system that includes not only visual biofouling estimation, but also control and localization of an underwater vehicle. 
A research project named HEKTOR (Heterogeneous autonomous robotic system in viticulture and mariculture) aimed to fill this knowledge gap and offer a solution that enables efficient coordination among heterogeneous autonomous robots, as can be seen in Figure \ref{fig:hektorSchema}. 
More information about the project can be found in \cite{hektorOverview1, hektorOverview2}.

One task that can, or better said should be automated, is the inspection of fish cage nets in aquaculture.
The net gradually accumulates biofouling which has many negative consequences. 
The main problem is the reduction in available area for clean water to flow through which causes the water inside the pen to become less oxygenated and more fouled, ultimately resulting in increased fish sickness and death rate.
Accompanying problems include the addition of extra mass to the pen structure, thus placing stress on the mooring ropes, damaging the net and causing a need for reknitting it. 
Periodic visual inspections by divers are currently necessary to assess the condition of the pens and determine the appropriate timing for cleaning.
Using an ROV or an AUV to perform said visual inspections would reduce the need for divers, helping to create a more streamlined, autonomous, and risk-averse inspection process. 

The proposed automated underwater fish cage inspection process includes an autonomous surface vehicle (ASV) working in cooperation with an ROV as shown in Figure~\ref{fig:hektorSchema}. 
The camera feed is streamed from the ROV, while biofouling estimation, control, and underwater algorithms are run on the ASV's onboard computer. 
The inspection process is envisioned to be split into two tasks. 
The first task is processing the footage obtained from a vehicle filming underwater to estimate the amount of biofouling present on the net.
The second task is developing an autonomous control algorithm to maneuver an ROV around the pens.

The main research goal was to develop an image processing algorithm that could be combined with the control algorithm in order to accurately assess the amount of biofouling accumulated on the nets.
This research contributes in several key areas.
Firstly, a neural network is successfully utilized for accurately segmenting underwater images of fish pens, enabling precise identification of pen structures. 
The architecture used for AI segmentation is the popularly employed UNet \cite{UNet}, explained in more detail later.
Secondly, the research explores and tests the feasibility of retrofitting an ROV with an underwater transponder to achieve precise localization. 
This modification improves upon its manual operation while also enabling autonomous missions, enhancing both its operational ease and versatility.
Thirdly, the paper proposes a technique for estimating the extent of biofouling buildup on a given pen. 
By combining image segmentation with localization, the technique provides a valuable means of quantifying the level of biofouling on fish pens.
Lastly, the proposed biofouling estimation technique is successfully tested in controlled and repeatable experimental scenarios. 
Collectively, these contributions improve fish pen biofouling analysis, and highlight potential advancements in the field of underwater robotics and mariculture practices.
\section{Equipment} \label{sec:Equipment}
\subsection{ASV Korkyra} \label{subsec: ASV}
\begin{figure}[!h]
\includegraphics[width=\linewidth]{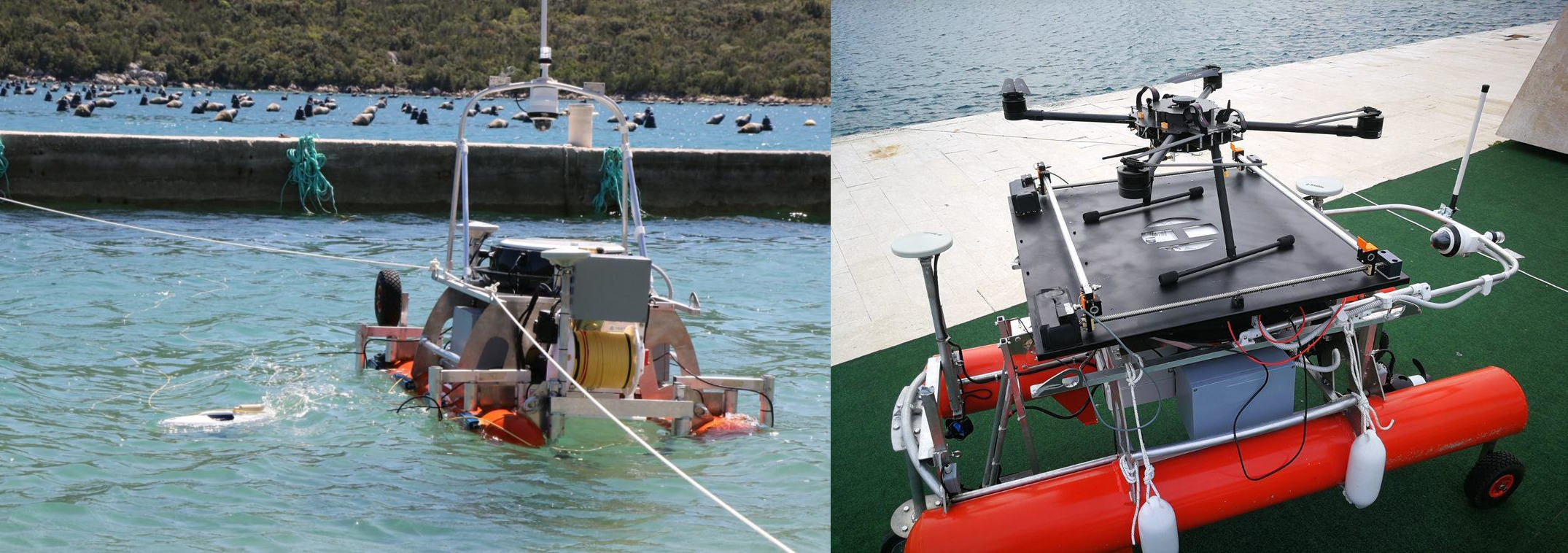}
\caption{Autonomous surface vehicle Korkyra: \textbf{left}---TMS mounted, \textbf{right}---LP mounted onto the~ASV.}
\label{fig:KorkyraShot}
\end{figure}

A custom-made aluminum catamaran named Korkyra was developed to function as a versatile remote-controlled or autonomous surface vehicle as a part of HEKTOR, as shown in Figure \ref{fig:KorkyraShot}. 
This specially designed catamaran boasts several key features, one of which is a landing platform dedicated to accommodating a lightweight drone for aerial operations \cite{hektorLandingPlatform}. 
Additionally, it incorporates a docking and tether management system intended for seamless integration with an ROV, enabling underwater mission capabilities, \cite{hektorTether}.
It also features a robust metal frame that enables the mounting of diverse tools and sensors such as cameras, sonar, lidar,~etc.
A powerful onboard computer is used to support real-time control algorithms, data processing, and other computational requirements.
With its purpose-built design and advanced functionalities, Korkyra serves as a valuable asset for remote-controlled or autonomous operations.

\subsection{ROV and Acoustic Localization System} \label{subsec: ROV mod} 
Autonomous systems must be able to precisely estimate their own position relative to obstacles, structures, and other objects in order to navigate and operate effectively in cluttered environments such as fisheries.
An underwater acoustic positioning system is essential for autonomous underwater missions due to the absence of GPS and standard RF-based positioning systems in underwater environments. 
This technology plays a vital role in enabling autonomous underwater missions by providing reliable positioning information where traditional positioning systems cannot operate effectively \cite{AcousticPosSystem}.
By using acoustic signals, the positioning system enables accurate navigation, mapping, and control of underwater vehicles in real-time. 

\begin{figure}[!h]
\includegraphics[width=\linewidth]{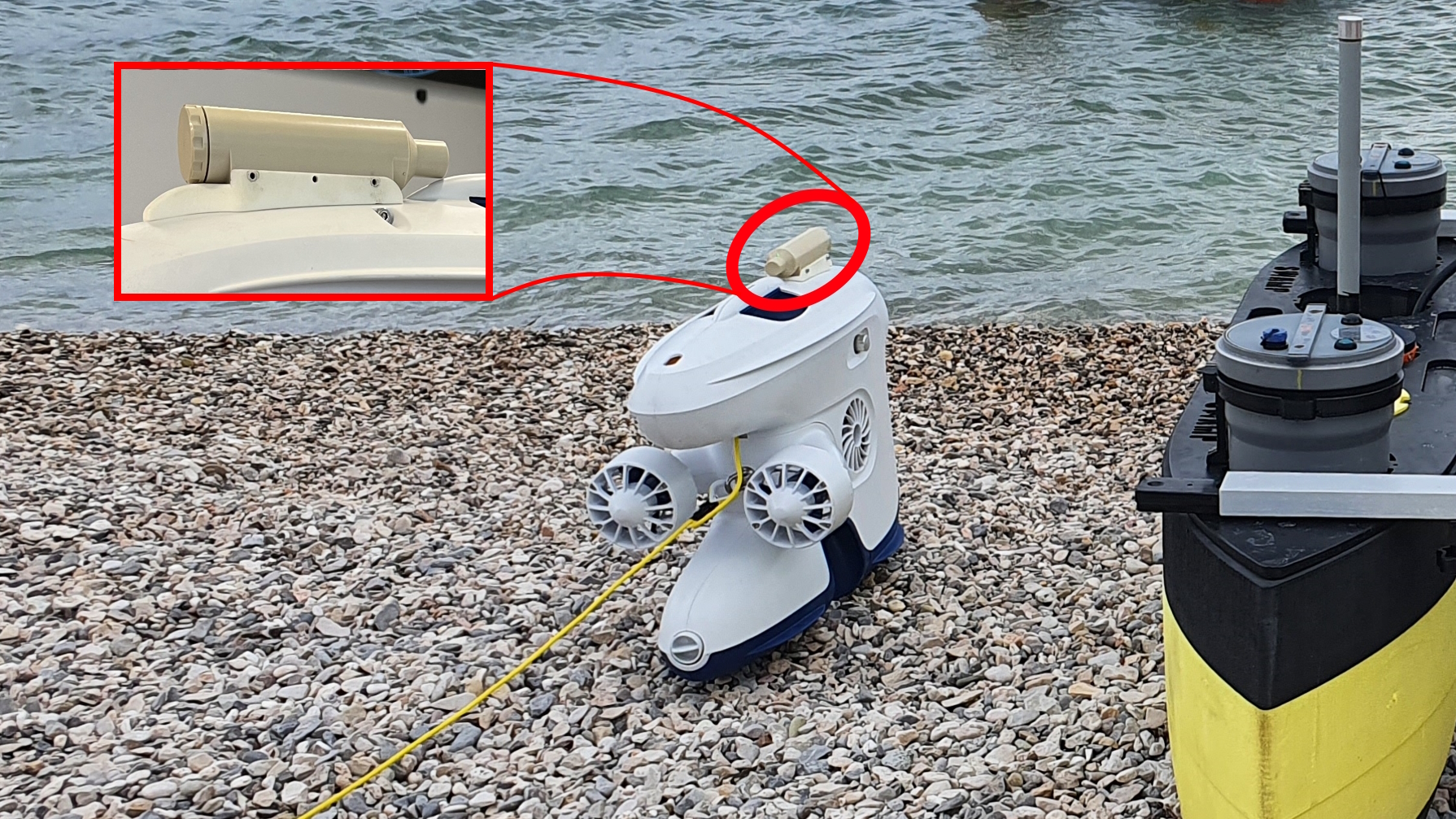}
\caption{The used Blueye Pro ROV and a closeup of the retrofitted WaterLinked Underwater GPS G2 transponder.}
\label{fig:ROVandUWGPS}
\end{figure}

A commercially available ROV was acquired from Blueye, a Norwegian company specializing in underwater technology. The ROV was mounted with a transponder belonging to an SBL acoustic underwater positioning system, as can be seen in Figure \ref{fig:ROVandUWGPS}.
The SBL acoustic localization system is the Underwater GPS G2 acquired from WaterLinked, also a Norwegian company specializing in acoustic subsea communication and positioning systems.
An L-shaped fixed position configuration with 4 transceivers placed at various depths, and 1 transponder mounted on the ROV was precise enough for use in a controlled test inspection scenario.
A secondary, but perhaps more realistic rectangular configuration with the 4 transceivers mounted on an ASV was also tested and provided good results in rough weather conditions.
The depth reading was taken directly from the ROV because the sensor readings were far more precise than those received from the SBL transponder.

\section{Methodology} \label{sec: Methodology}

\subsection{Dataset Collection and Labeling} \label{subsec: Data gathering and labeling}
\begin{figure}[!ht]
\includegraphics[width=\linewidth]{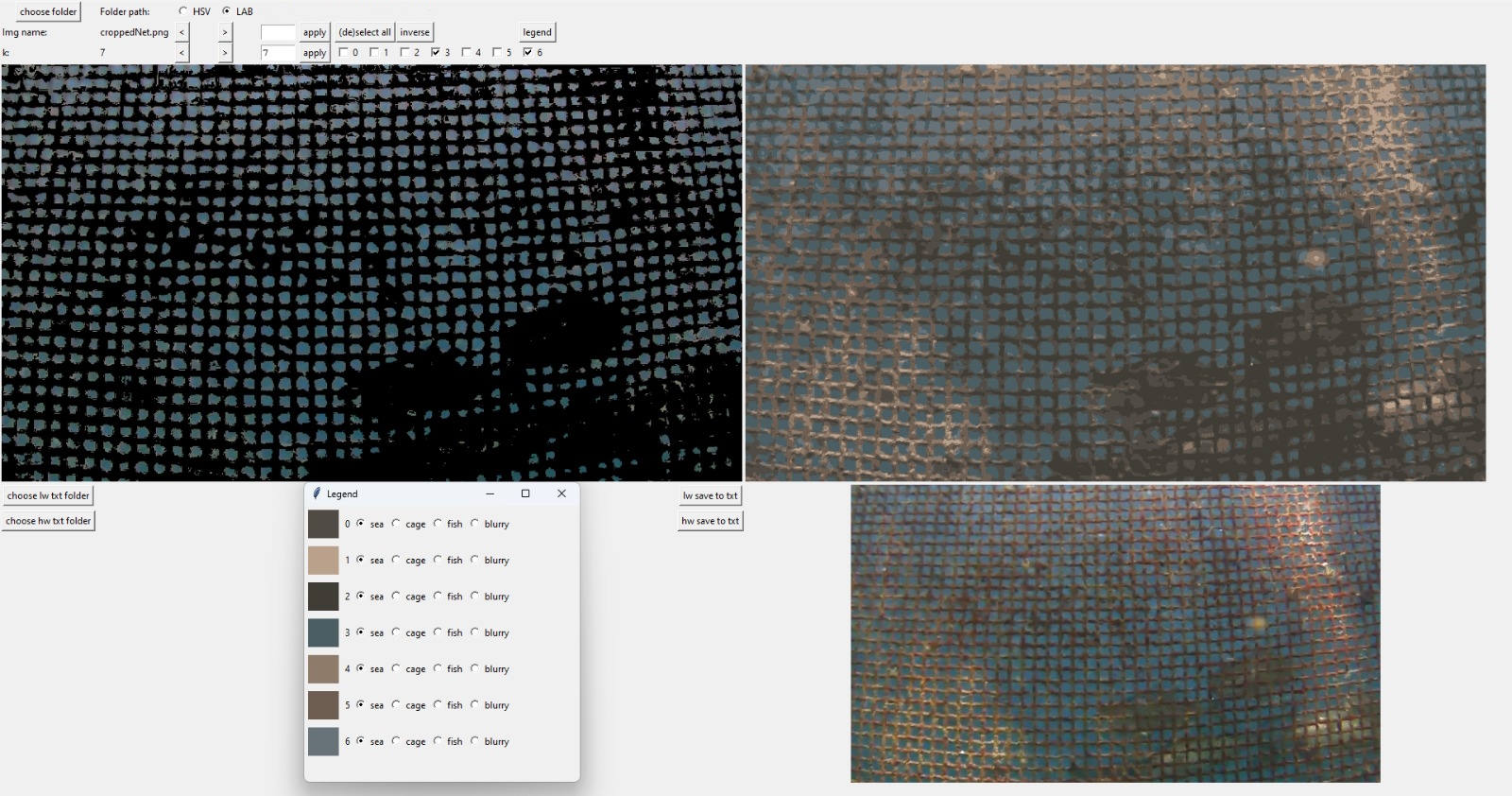}
\caption{Screenshot of the labeling tool developed for easier dataset creating. The image in the top left is the original image with clusters of pixels turned on or off. The image in the top right replaces pixels from the original image with their respective centroids resulting from K-Means clustering. It is possible to choose the color space of the image, change the K hyperparameter, and turn the pixels on and off using the toolbar above the images. The legend pop-up window in the bottom left is used for turning the grouped pixels ``on or off'' and assigning labels. The color squares represent the resulting centroids of K-Means clustering.}
\label{fig:labelTool}
\end{figure}

The research involved collecting underwater video footage of fish cages in the Adriatic Sea during the summers of 2020 and 2021 to study biofouling buildup. 
This footage, showcasing cages in various states of fouling, was used to create a dataset of nearly 4,000 images. 
Additional footage was collected in a controlled seawater pool in 2022, adding around 1,000 more images.
To analyze the buildup, the research focused on image segmentation, requiring labeled images of the cages. 
A labeling tool was developed to streamline this process by using the K-Means clustering algorithm to group pixels of similar colors. 
This approach effectively handled large image volumes, simplifying images down to key colors without losing structural detail. 
The tool allowed operators to label images efficiently by selecting clusters and assigning them to categories such as ``sea'', ``cage'', ``fish'', and ``blurry'' as shown in Figure \ref{fig:labelTool}.

\subsection{Biofouling Estimation Framework} \label{subsec: Biofouling estimation framework}
\begin{figure}[!ht]
\includegraphics[width=\linewidth]{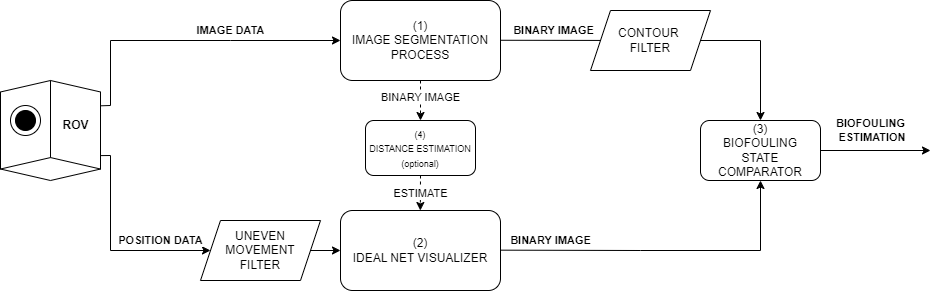}
\caption{Schematic of the used biofouling estimation framework.}
\label{fig:estimFrameworkMaster}
\end{figure}

The modular biofouling estimation framework proposed herein is depicted in Figure~\ref{fig:estimFrameworkMaster}.
The framework comprises several interconnected nodes with each node serving a specific purpose.
The framework and the nodes came about as a product of developing image processing algorithms and the ROV control loop in the Robotic Operating System (ROS).
Node (1), shown in Figure \ref{fig:estimFrameworkMaster}, is responsible for executing the image segmentation process, separating the fish pen structure and its net from the background. 
Node (2) combines the segmented data with the distance information to reconstruct how a clean net would appear at that specific filming distance.
Node (3) compares the two binary images, one of the ideal net state and one of the current state, and the result quantifies the extent of biofouling coverage on the net's surface.
Node (4) implements pose (distance) estimation from a single camera if the distance from the filmed net is not known from a 3D map of the environment or some other source.

\begin{figure}[!ht]
\includegraphics[width=\linewidth]{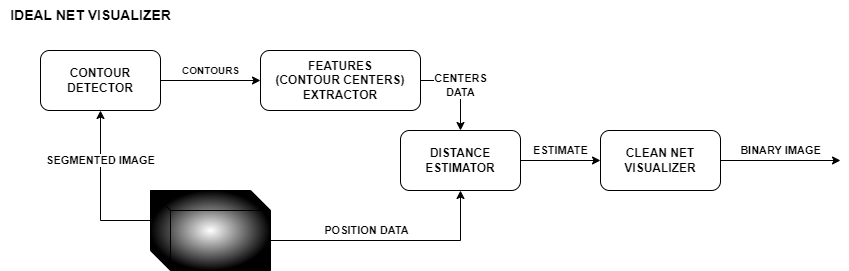}
\caption{Ideal net visualizer part of the framework shown in more detail as it is implemented.}
\label{fig:estimFrameworkDetail}
\end{figure}

A particular implementation of the framework developed during research calculates an estimated distance from the net by determining the approximate distance between each center of the small squares in the net, shown in more detail in Figure \ref{fig:estimFrameworkDetail}. 
The selection of centers as good features in the biofouling estimation process was based on their inherent stability.
The squares gradually reduce in size as biofouling builds up on the net, however the position of the center remains constant which makes the centers robust features for detection.
By having knowledge of the distance in pixels for specific features on an object in an image, as well as the corresponding real-life distances, along with information about the camera sensor used for capturing the picture such as the physical size of the sensor and its focal length, it is possible to estimate the distance from the observed object to the camera. 
This allows for a reasonable estimation of the distance from the camera to the observed fish pen.
This method of distance estimation can be classified as a geometric approach to the problem, as opposed to using a more time-consuming and computationally heavier deep-learning approach. 

\subsubsection{Biofouling Buildup Quantification} \label{subsubsec: Biofouling Quantification}
The net is filmed with as uniform a movement as possible at a fixed distance from the net.
The ROVs movement speed was such that filming at 1Hz was frequent enough to capture the entire net area.
Uniform movement of the ROV ensures that each segment of the net is filmed for roughly the same amount of time.
The entire footage can be fed into the framework, and the amount of biofouling for the filmed area would be the average of all of the estimations.

\subsubsection{Image Segmentation Node} \label{subsubsec: Image Processing Node}
The labeled data from the image labeling process was initially used to train a logistic regression model, which aimed to classify pixels as belonging to the fish cage structure or the background (sea and fish) based on color. This model, trained on an 80-20\% train-test split, showed limitations in handling variations like lighting conditions and image blurriness, leading to its abandonment \cite{hektorCageInspection2}. 
A more robust approach using neural networks, specifically the UNet architecture, was then adopted for image segmentation. 
UNet, known for its U-shaped design, effectively captures both high-level context and fine details, producing accurate pixel-wise segmentation masks \cite{UNet1st, UNet}.
The model was trained on randomly selected images from the dataset, using the same 80-20\% split, with only basic sharpening applied to images before segmentation. 
The goal was to accurately distinguish between the fish pen net and the surrounding background.

\subsection{Closed-Loop ROV Control System} \label{subsec: ROV control loop}
\begin{figure}[!ht]
\includegraphics[width=\linewidth]{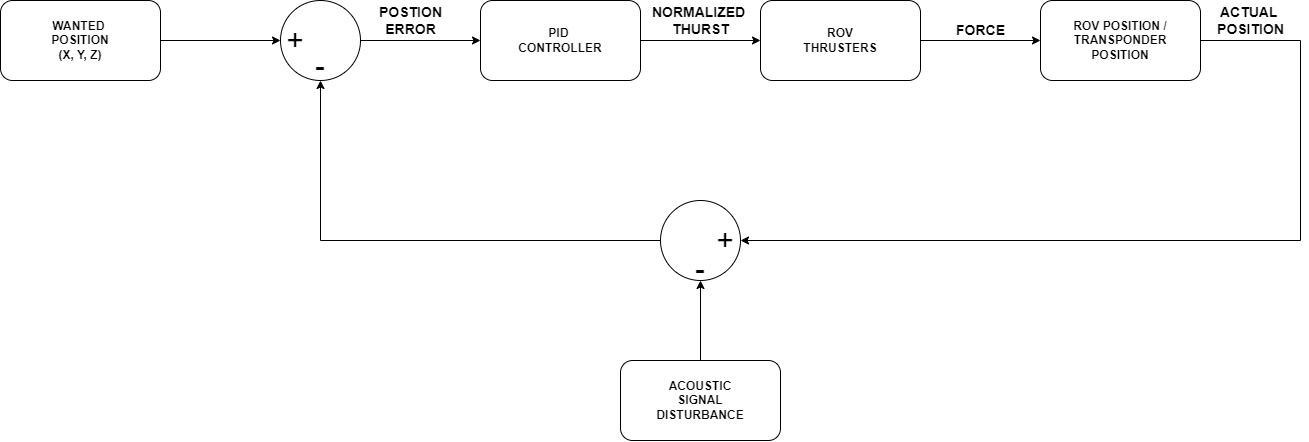}
\caption{Schematic depicting the ROV control loop.}
\label{fig:controlLoop}
\end{figure}

\begin{figure}[!ht]
\includegraphics[width=\linewidth]{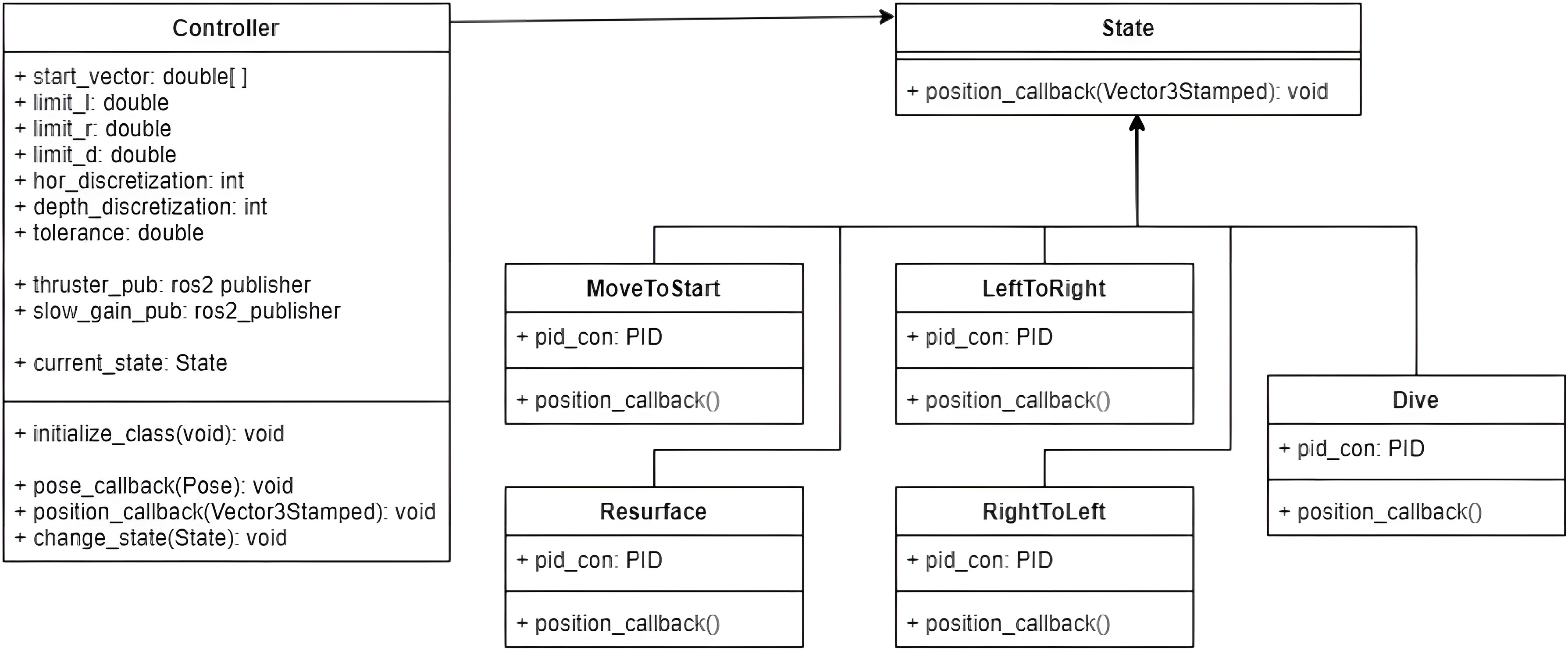}
\caption{Schematic showing the implemented control loop class diagram. The ``Controller'' class generates waypoints by taking into account the leftmost and rightmost possible values for the position, depth, and how many discrete points to generate along the horizontal and vertical axes. The outputs are values for thruster speed along two controlled degrees of freedom (surge and sway).}
\label{fig:classDiagram}
\end{figure}

A lawnmower pattern trajectory controller was designed to serve as a proof-of-concept to test the possibility of complete autonomous control for the modified ROV.
This test aimed to assess functionalities such as precise position estimation, efficient trajectory planning, and to validate the responsiveness of the ROV's control loop.
Each state in the control algorithm represents a specific action for the ROV, such as moving to a starting position, swaying, descending, or resurfacing to a predetermined end point. 
A general schematic for the control system can be seen in Figure \ref{fig:controlLoop}, and the different states of the controller can be seen in a UML class diagram of the controller implementation in Figure \ref{fig:classDiagram}.
The control system implemented for the ROV managed surge and heave motions.
Additionally, the ROV features built-in automatic heading maintenance, that is, it points to a constant direction during operation.
The position error is used as an input into a classic PID controller that generates thruster commands, so that the control system can be used for any 2-DOF ROV \cite{PIDControl}.

\section{Experimental Setup} \label{sec:Experimental Setup}
\begin{figure}[!h]
\includegraphics[width=\linewidth]{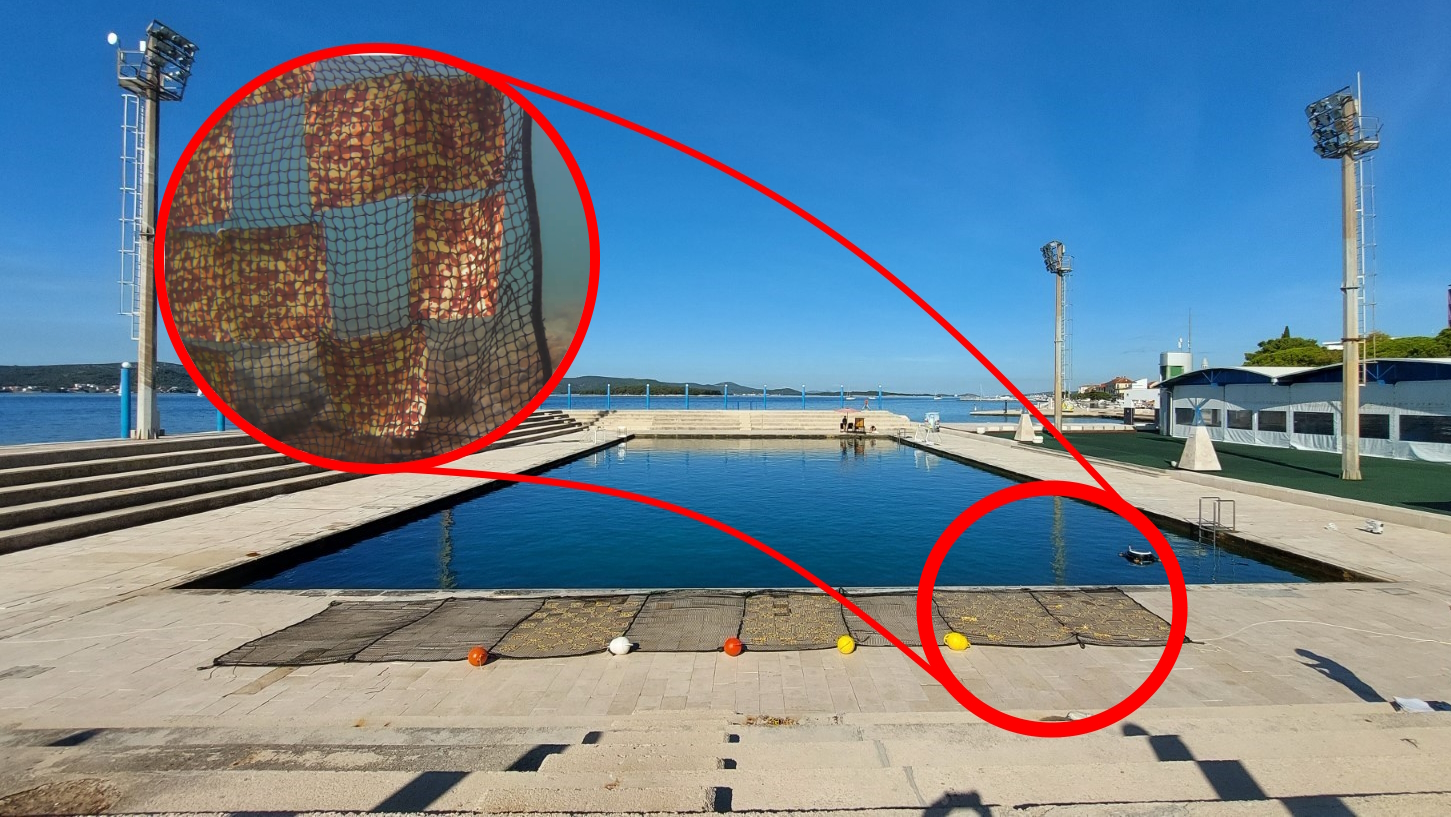}
\caption{The Olympic sea water pool used for testing, with the net dragged out in order to place patches for biofouling simulation. The affixed patches on the net can be seen close$-$up filmed during a mission in the red circle.}
\label{fig:poolAndPatch}
\end{figure}

Validation experiments for the biofouling estimation framework were conducted in a controlled seawater pool in Biograd, Croatia in late September and early October of 2022.
For the purpose of conducting these trials, a pen net was acquired from an industrial fish farm and deployed within the controlled environment of an Olympic-sized pool.
The dimension of the net is $\sim$14 m wide and $\sim$3 m high, so $\sim$42\,m$^2$ of area in total.
To simulate biofouling buildup in this scenario, camouflage-pattern colored square patches were strategically hand-placed onto the fish cage net as can be seen in Figure \ref{fig:poolAndPatch}.
The patches were 25cm x 25cm squares.
They were designed to mimic the visual appearance of underwater biological fouling using a brown--yellow color scheme. 
An increasing number of patches were added to the net in each iteration of the experiment.
The net was filmed once with no patches, once with patches covering 22\%, 33\%, and 44\%, and three times with patches covering 66\% of the net with greater distance to the net each time.
The ROV was autonomously controlled during the underwater missions. 
The implementation of autonomous control in the controlled ocean-floor pool environment not only facilitated more consistent data collection but also served to test the developed control algorithm.

\begin{figure}[!h]
\includegraphics[width=\linewidth]{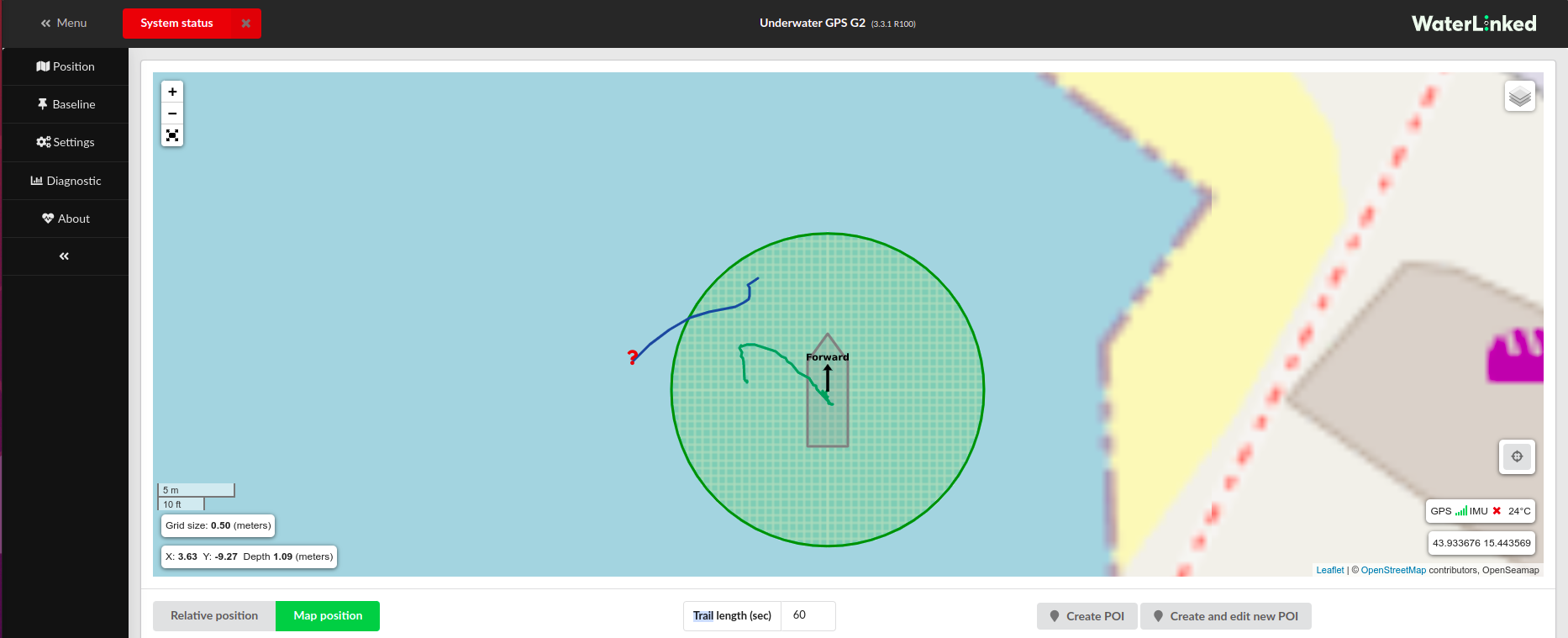}
\caption{Screenshot of the UWGPS GUI while the ROV is tethered to the ASV which also acts as the carrier for the short baseline transponder setup. Since the trajectory of the ROV is visualized, this software was used to roughly estimate the precision of the SBL setup.}
\label{fig:UWGPSWorking}
\end{figure}

A screenshot of the UWGPS system GUI can be seen in Figure \ref{fig:UWGPSWorking}, taken during one testing of the ASV+ROV combination in open sea.
The green line represents the ASV trajectory which should always be available as it has the UWGPS box mounted, while the blue line represents the trajectory of the free-moving transponder that is attached to the ROV in this case, and can travel out of the search range.
Due to limitations of working in a pool, a fixed baseline of transponders was positioned around the pool's edge.
Although this approach differed from the anticipated ASV + ROV combination, it provided a needed practical solution for achieving reliable positioning data during the experimental setup in the pool environment.

\section{Results} \label{sec:Results}
\subsection{Autonomous ROV Control Loop Results} \label{subsec: ROV Results}
\begin{figure}[!ht]
\includegraphics[width=\linewidth]{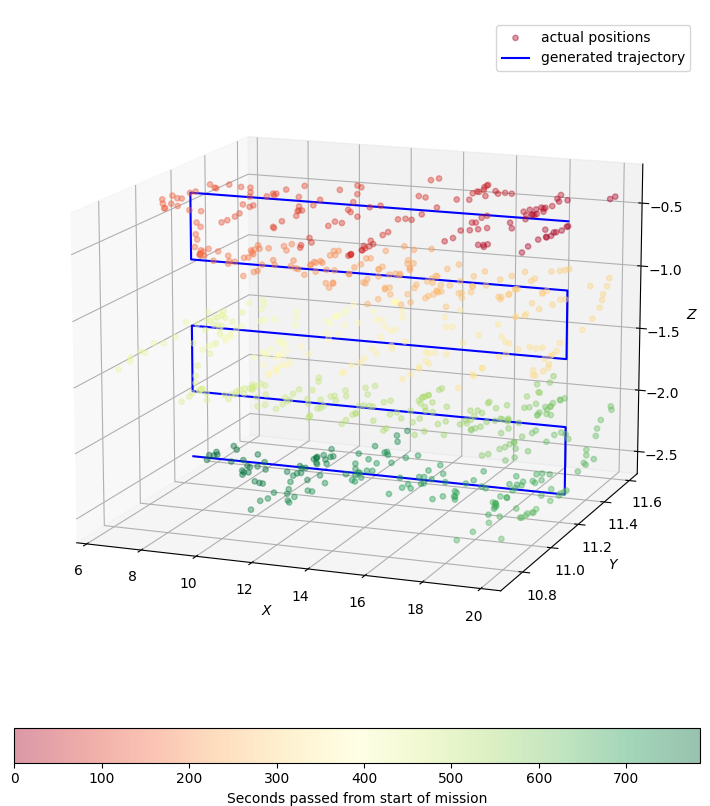}
\caption{3D plot of the ROV position. The blue line represents a perfect trajectory, the colored dots represent actual positions in time during one mission.}
\label{fig:3dMovementPlot}
\end{figure}

Each ROV mission took around 15 min to cover the aforementioned $42\, \text{m}^{2}$.
The 3D plot in Figure \ref{fig:3dMovementPlot} visualizes the ROV's movement during a mission, and its ideal trajectory generated by the control algorithm. 
The Y-axis in Figure \ref{fig:3dMovementPlot} correlates to the distance from the ROV to the net since the net was strung out straight.
The controller was implemented with this assumption, so it tried to keep a constant Y-coordinate throughout the mission.
The position is plotted every second. 

Having the ROV's surge speed be between $\sim$0.1 m/s and $\sim$0.2 m/s has shown to be a good compromise between the quality of footage regarding image sharpness, and the duration of a mission. 

\subsection{Image Labeling Results} \label{subsec: Labeling Results}
The images picked for the dataset were represented in the LAB color space. 
The LAB color space separates the luminance (L) channel from the color information.
During the image labeling process, only the central part of the full HD image was considered due to concerns related to camera distortion and overall poor image quality near the edges. 
Images exhibiting labeling errors, low image quality, or inconsistencies were identified and removed from the training dataset. 
Table \ref{tab:images per year} shows the distribution of images in the dataset.

\begin{table}[!ht]
    \caption{\label{tab:images per year}Table showing how many images per filming year were labeled in total, and how many of the labeled images ended up in the training/validation dataset for the neural network.}
    \setlength{\tabcolsep}{2.35mm}
    \resizebox{\linewidth}{!}{%
    \begin{tabular}{ c  c  c c }
     \toprule
    \textbf{Footage year} & \textbf{Location} &\textbf{Labeled imagess} & \textbf{Images in dataset} \\ 
     \midrule
     2020 & Ugljan & 938 & 261 \\  
      \midrule
     2021 & Ugljan & 405 & 197  \\
      \midrule
     2022 & Biograd & 919 & 694 \\
     \bottomrule
\end{tabular}}  
\end{table}
\vspace{-6pt}

\subsection{UNet Architecture Segmentation Results} \label{subsec: UNet results}
As mentioned earlier, in order to train the UNet model the standard 80--20\%
 train--test split was used, meaning that the model was trained on 80\% of available annotated images in the dataset, and the results of the training steps were validated on a randomly selected 20\% of images in the dataset that are unseen during training.
The training process was stopped once no more discernible improvement was shown from one iteration to the next.

\begin{figure}[!ht]
\includegraphics[width=\linewidth]{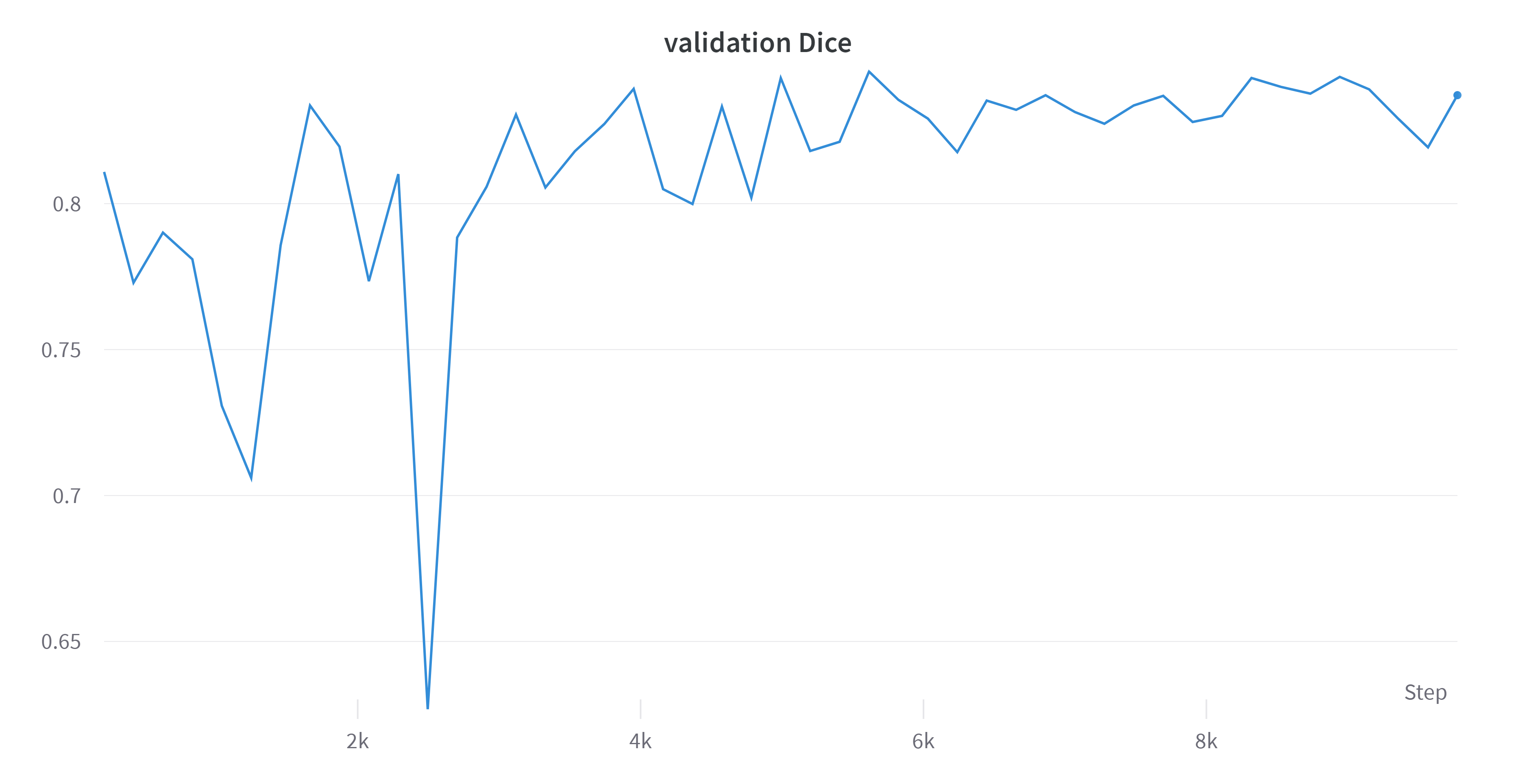}
\caption{Plot showing the popular Dice score used in image segmentation analysis and its change during the training.}
\label{fig:validationPlot}
\end{figure}

\begin{figure}[!ht]
\includegraphics[width=\linewidth]{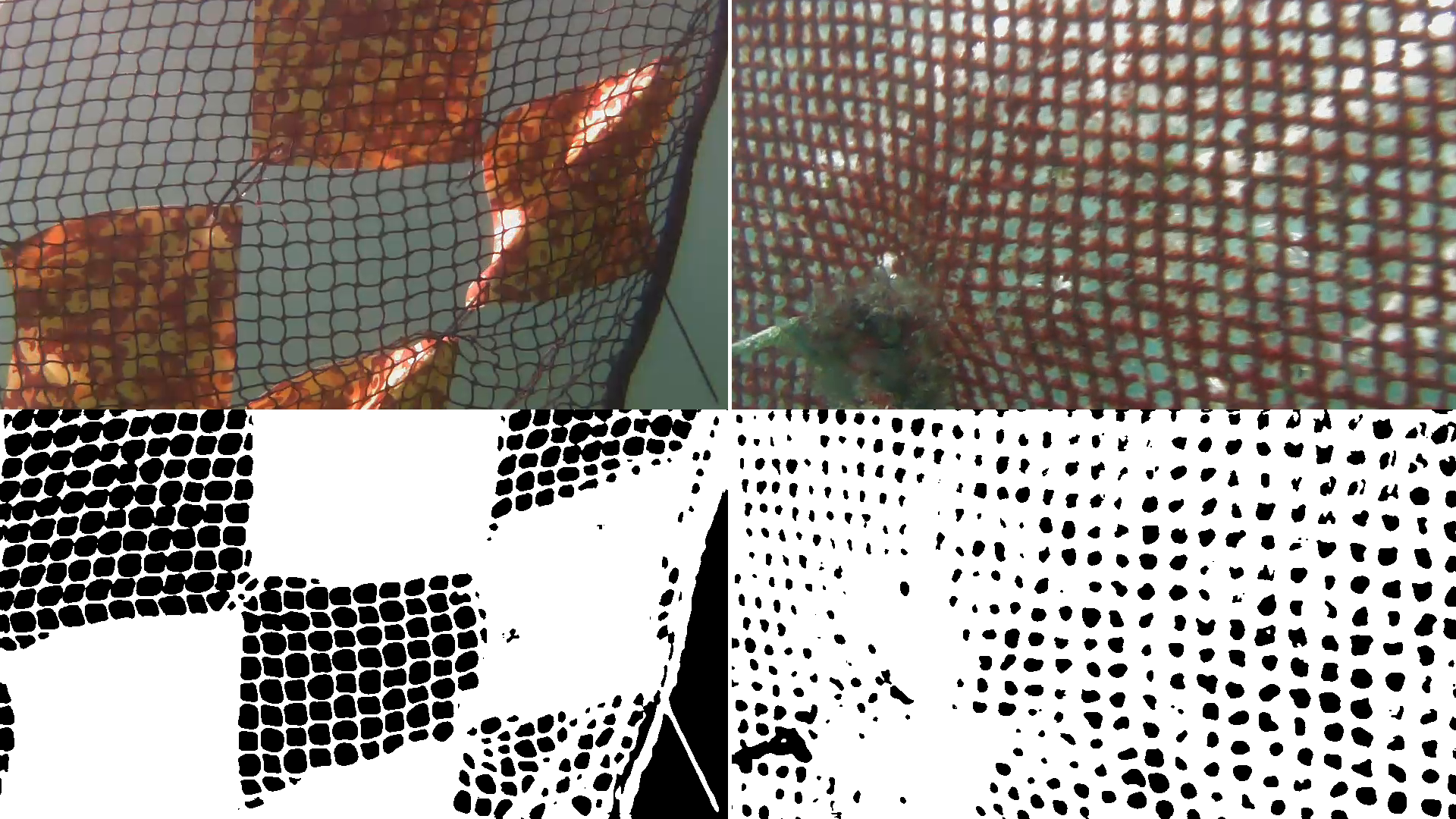}
\caption{
Result of predictions made by a trained UNet neural network segmentation model for: \textbf{left}---an image taken in the controlled conditions in Biograd, \textbf{right}---an image taken at a real fishery near Ugljan.}
\label{fig:segmented_joined}
\end{figure}

Dice score is a common metric used for scoring the performance of image segmentation models, ranging from $0.0$ to $1.0$ (a higher value is better). 
It measures the similarity or overlap between the predicted segmentation mask and the ground truth segmentation mask \cite{dice}.
The highest Dice score achieved on the validation set of images was 0.8434 during the 9th training epoch, and the weights calculated to achieve this coefficient were saved to be used later on.
The Dice score can be seen changing during the training process in Figure \ref{fig:validationPlot}.
A visualization of how the trained UNet model successfully segments the images can be seen in Figure \ref{fig:segmented_joined}.

\vspace{-6pt}

\subsection{Biofouling Estimation Results} \label{subsec: Biofouling Estimation results}
As previously mentioned, an industrial fishery provided a fish pen net for the experimental setup. 
Square patches were affixed onto the net in a series of iterations to simulate biofouling.
The patches were incrementally added in four stages, progressively covering larger areas of the net. 
All of missions had the ROV at around 1 m away from the net being filmed, except a repeated mission at 66\% where the distance was purposefully increased to around 1.5 m.
To establish a benchmark result, the biofouling estimation algorithm was initially applied without the implementation of any filtering methods.
All of the recorded footage in a filming session was only cropped around the center of the frame and fed into the estimation algorithm.
Table \ref{tab:error percentages} shows the benchmark estimated percentages.

\begin{table}[!ht]
\setlength{\tabcolsep}{18.95mm}
\caption{\label{tab:error percentages}Table showing the actual simulated biofouling percentage, and the estimated biofouling percentage generated by the estimation framework.}
\resizebox{\linewidth}{!}{%
\begin{tabular}{ll}
\toprule
\textbf{Actual Biofouling} & \textbf{Estimated Biofouling} \\ \midrule
22.00\%                       & 16.00\%                      \\ \midrule
33.00\%                       & 32.19\%                       \\ \midrule
44.00\%                       & 41.02\%                      \\ \midrule
66.00\%                       & 65.48\%                         \\ \bottomrule
\end{tabular}}

\end{table}

\subsubsection{Framework Filters Results Using UNet for Semantic Segmentation} \label{subsubsec: Framework filters UNet}
\begin{figure}[!ht]
\includegraphics[width=\linewidth]{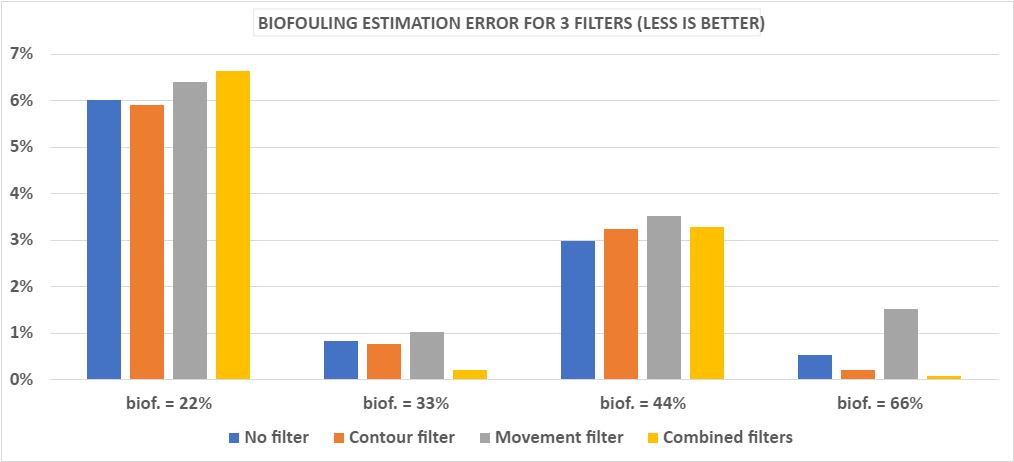}
\caption{Bar plot showing the error in the estimated percentage of biofouling.}
\label{fig:PercentageErrorPlot}
\end{figure}

The two filter methods implemented were the exclusion of footage during non-uniform movement, and the contour filtering based on size and shape of detected contours.
The difference to the estimated percentage that the filters made can be seen in Figure \ref{fig:PercentageErrorPlot}.
The apparently small change is due to the fact that the autonomous filming worked well in a sense that the ROV's speed was consistent throughout the mission and the angle of the filming was good.
Each area of the net is filmed for around the same amount of time, and the neural network semantic segmentation model performed well, so the benchmark result without filtering was close to correct from the start.
When the filming conditions are not perfect, such as in the repeated 66\% coverage scenario tested in Biograd, then the filters help out.
The filming conditions were purposefully worsened by having a greater distance from the ROV to the net during filming.
The combination of both filters reduced the estimation error by 1.75\% in total, as can be seen in Table \ref{tab:bad filming conditions filters}.

\begin{table}[!ht]
\caption{\label{tab:bad filming conditions filters}Table showing the actual simulated biofouling percentage, and the estimated biofouling percentage generated by the estimation framework with the combinations of the footage and computer vision filters turned on.}
\setlength{\tabcolsep}{8.95mm}
\resizebox{\linewidth}{!}{%
\begin{tabular}{lcc}
\toprule
\textbf{} & \multicolumn{1}{l}{\textbf{Value}} & \multicolumn{1}{l}{\textbf{Biofouling Estim. Error}} \\ \hline
Actual biofouling            & 66.00\%          & /             \\ \midrule
No filter estim.             & 48.04\%       & 17.96\%       \\ \midrule
Contour filter estim.        & 49.44\%       & 16.56\%       \\ \midrule
Movement filter estim.       & 48.98\%       & 17.02\%       \\ \midrule
Combined filter estim.  & 49.79\%       & 16.21\%       \\ \bottomrule
\end{tabular}}
\end{table}

\subsubsection{Effect of Filming Conditions} \label{subsubsec: Filming conditions}
An issue arose during a control test where the ROV's heading angle was fixed, and a clean net with no biofouling patches was filmed. 
Despite the net being clean, the estimation incorrectly reported 32.37\% biofouling due to several factors: the net's slight convexity and waviness (as it was not tightly strung), the ROV's heading potentially drifting due to an imperfectly calibrated compass, and filming from a tilted angle. 
Furthermore, one of the factors during filming is the time of day and the position of the Sun.
The used Blueye Pro ROV does not have an HDR camera, meaning that an overexposure of one part of the camera sensor to light ruins the rest of the image.
This effect poses a problem when filming near the surface.
Filming missions should be planned accordingly, holding them early in the morning or late afternoon, or filming the cages with the Sun behind the camera.
In addition, it goes without saying that the filming distance greatly impacts image quality and the estimation process as a whole.
Filming should be done at a distance where the the net can be clearly separated from the background, i.e., the edges should be sharp and easily discernible.
As mentioned earlier in Section \ref{subsec: Labeling Results}, the sea floor could have a big impact on the computer vision component of the framework.
Having the sea floor visible increases the already difficult challenge of accurate semantic image segmentation.
Luckily, the scenario is unlikely because fish farms should be situated 3km away from shore and have 50 meters of depth available to limit the environmental impact, as mentioned in \cite{holmer2010environmental, offshoreFishFarm}.

\subsubsection{Choosing the UNet Architecture Instead of Logistical Regression} \label{subsubsec: Choosing UNet}
\begin{figure}[!ht]
\includegraphics[width=\linewidth]{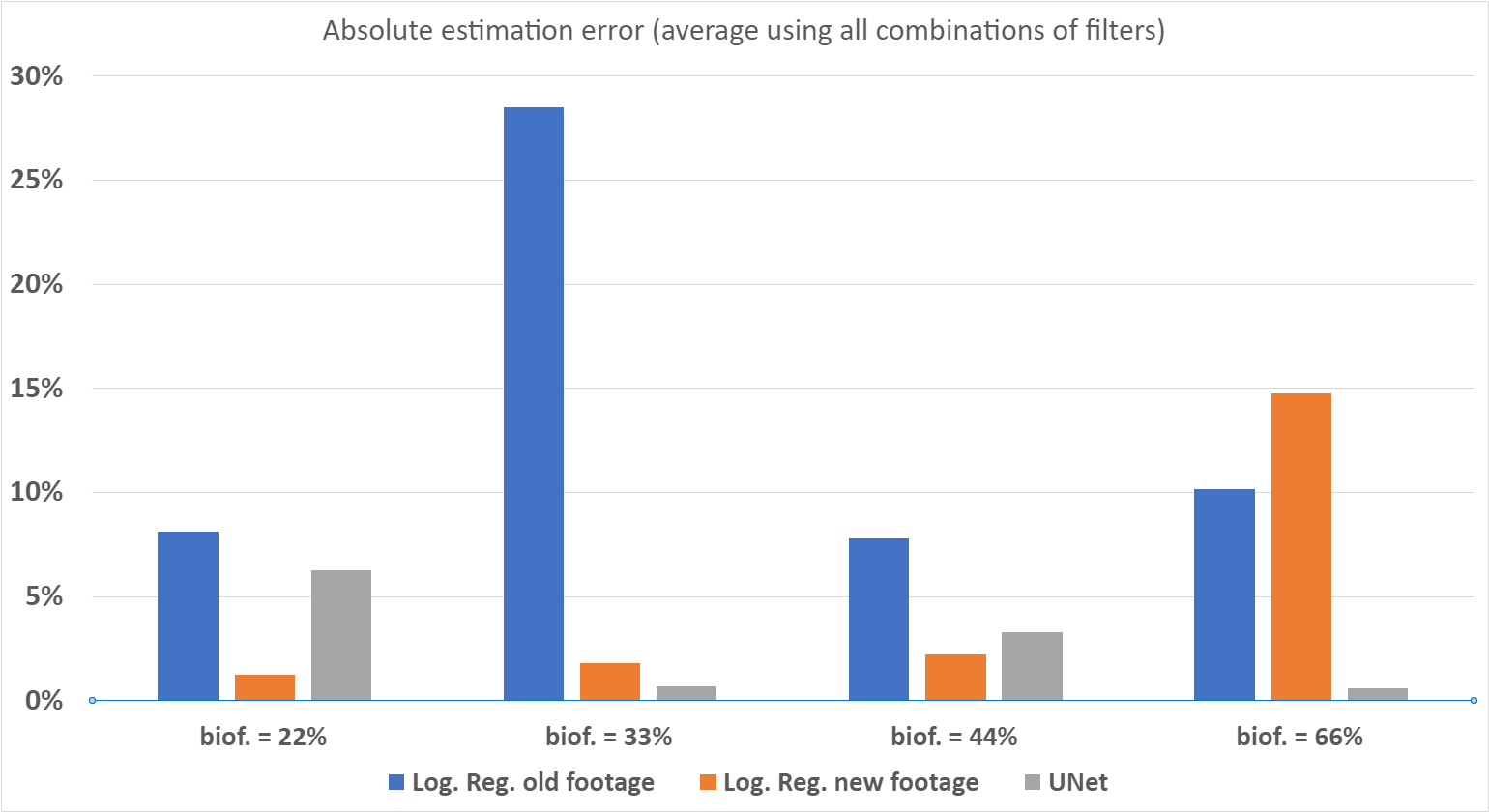}
\caption{Bar plot showing the average absolute error in the estimated percentage of biofouling for logistical regression models trained on just old (blue bar) and combined (orange bar) data, and UNet.}
\label{fig:LogRegBarPlot}
\end{figure}

It is important to note that the results seen so far were achieved using the trained UNet architecture model for semantic image segmentation.
While logistical regression initially seemed promising for semantic segmentation (previous research done in \cite{hektorCageInspection2}), its limitations became evident during testing in Biograd.
It quickly became apparent that the method is not robust enough.
Small changes in lighting conditions completely threw off the segmentation which then produced unusable results.
Adding the footage captured in Biograd into the training dataset does improve estimation results, but at the same time it degrades the quality of segmentation from footage captured in previous years.
Still, the biofouling estimation algorithm was run with both the old and new (added footage from Biograd) logistical regression models, and the results can be seen in Table \ref{tab:log reg table}.
All of the estimated percentages in the table are generated by the framework without using any of the filters mentioned before.
It is apparent that using logistical regression not trained on new footage produces much more inaccurate results than the other two models in Figure \ref{fig:LogRegBarPlot}.
Training a logistical regression model with new footage does improve performance, but at the cost of overtraining which can be seen by poor performance for the highest biofouling scenario.
The results for that particular test are worse because the filming conditions are different in a sense that the previous three filming missions were carriedo ut on Wednesday afternoon, and the 66\% biofouling coverage filming was done the next day on Thursday morning.

\begin{table}[!ht]

\caption{\label{tab:log reg table}Table showing the average estimate of biofouling percentage when using the footage filters, for each semantic segmentation model.}

\setlength{\tabcolsep}{5.95mm}
\resizebox{\linewidth}{!}{%
\begin{tabular}{cccc}
\toprule
\multicolumn{1}{l}{\textbf{Actual Biofouling}} &
  \textbf{\begin{tabular}[c]{@{}c@{}}Log. Reg. Estim.\\ (Old Footage)\end{tabular}} &
  \textbf{\begin{tabular}[c]{@{}c@{}}Log. Reg. Estim.\\ (New Footage)\end{tabular}} &
  \multicolumn{1}{l}{\textbf{UNet}} \\ \midrule
22\% & 29.96\% & 20.99\% & 16.00\% \\ \midrule
33\% & 50.12\% & 34.63\% & 32.19\% \\ \midrule
44\% & 53.49\% & 41.73\% & 41.02\% \\ \midrule
66\% & 52.45\% & 50.05\% & 65.48\% \\ \midrule
\end{tabular}}
\end{table}

To sum up, using a robust UNet model for the image segmentation node in the biofouling framework produced an average absolute biofouling estimation error of just 2.54\%, as can be seen in Figure \ref{fig:LogRegBarPlot}.
The image segmentation is the most computationally heavy task of the framework, and the time taken to segment an image using a trained UNet model depends on the power of the onboard computer in the ASV if the estimation is to be done in real-time, or the workstation if the estimation is to be carried out after filming is complete.
Using a dedicated GPU to run the UNet model is heavily recommended because it reduces the processing time of a 960 $\times$ 540 resolution image from a few seconds using a CPU, to the millisecond order of magnitude using a GPU. 
Furthermore, using UNet for segmentation means that potentially obtaining a diverse range of footage from various fisheries filmed under different conditions could only enhance the model's robustness and performance, whereas opting for simpler models like logistical regression would lead to overtraining when faced with varying scenarios.
\section{Conclusion} \label{sec: Conclusion}
In conclusion, the main contributions of this research were: (1) development of a labeling tool in order to create a curated dataset of labeled underwater HD images of fish pens, (2) development and implementation of a framework successfully used for estimating the amount of biofouling on the nets of fish pens, that incorporates a trained AI neural network model used for the task of semantic segmentation of underwater images of fish pens, and (3) development of an autonomous closed-loop control system using the available SDK for the ROV and the available localization data supplied by the retrofitted underwater positioning~system.

The control loop algorithm successfully controlled the ROV in a pool setting using point-to-point navigation.
Although the scenario in which the experiment was conducted is not an exact replica of the actual conditions in a fish farm, it still demonstrates the possibility of using the ROV as an autonomous vehicle to perform inspection missions.
To achieve localization in a complex environment the control algorithm could be modified to include a three-dimensional map of the farm and also fuse SONAR measurements if such a sensor would be mounted onto the ASV, together with UWGPS and live camera footage.
The labeling tool made it possible to accurately segment and semantically label the images of fish pens at around 30 s or less per image, thus allowing us to create a dataset of more than a thousand images within a satisfactory time frame. 
Perhaps most important, the implementation of the proposed framework which was tested in a controlled environment proved to be a success with the absolute value of the estimation error roughly being 2.5\%.
It is also worth noting that carrying out the inspection mission in one take while keeping the velocity of the ROV almost constant throughout, without much backtracking or spending too much time filming one area in relation to the rest of the net, produces an accurate estimation.
Precisely determining filming positions to keep the overlap of footage fed to the estimation framework at a minimum might not be cost-effective to develop.
Due to the nature of the current inspection process which involves divers estimating the state of the net, a ``good enough'' estimation is satisfactory.

For future endeavors, the developed framework should be tested at an industrial fishery using the original idea of pairing the ASV Korkyra with an ROV.
This field testing would aim to validate the framework's effectiveness in a challenging environment like an industrial fish farming operation.
Furthermore, the previously developed proprietary tether management system \cite{hektorTether} that also relies on localization of the ROV should be integrated physically onto the ASV.
Lastly, the project also envisions an air surveillance aspect using a light autonomous drone that could take off and land from the ASV \cite{hektorLandingPlatform}. 
The combined heterogeneous system of robots should be tested in a real-world scenario in the future.

\bibliographystyle{IEEEtran}
\bibliography{references}
\end{document}